\documentclass[10pt,twocolumn,letterpaper]{article}

\usepackage[table]{xcolor}
\usepackage{iccv}
\usepackage{times}
\usepackage{epsfig}
\usepackage{graphicx}
\usepackage{amsmath}
\usepackage{amssymb}
\usepackage{multirow}

\usepackage[pagebackref=true,breaklinks=true,letterpaper=true,colorlinks,bookmarks=false]{hyperref}

\iftrue
\newcommand{\davide}[1]{\textcolor{blue}{\bf [DAVIDE: #1]}}

\newcommand{\joe}[1]{\textcolor{magenta}{\bf [JOE: #1]}}
\newcommand{\todo}[1]{\textcolor{red}{\bf [TODO: #1]}}

\else
\newcommand{\davide}[1]{}
\newcommand{\rahul}[1]{}
\newcommand{\bing}[1]{}
\newcommand{\joe}[1]{}
\newcommand{\todo}[1]{}

\fi

\iccvfinalcopy %

\def\iccvPaperID{6483} %

\ificcvfinal\fi
\begin{document}

\title{Action recognition with spatial-temporal discriminative filter banks}

\author{Brais Martinez, Davide Modolo, Yuanjun Xiong, Joseph Tighe\\
Amazon\\
{\tt\small braisa,dmodolo,yuanjx,tighej@amazon.com}
}

\maketitle

\begin{abstract}
Action recognition has seen a dramatic performance improvement in the last few years. Most of the current state-of-the-art literature either aims at improving performance through changes to the backbone CNN network, or they explore different trade-offs between computational efficiency and performance, again through altering the backbone network. However, almost all of these works maintain the same last layers of the network, which simply consist of a global average pooling followed by a fully connected layer. In this work we focus on how to improve the representation capacity of the network, but rather than altering the backbone, we focus on improving the last layers of the network, where changes have low impact in terms of computational cost. In particular, we show that current architectures have poor sensitivity to finer details and we exploit recent advances in the fine-grained recognition literature to improve our model in this aspect. With the proposed approach, we obtain state-of-the-art performance on Kinetics-400 and Something-Something-V1, the two major large-scale action recognition benchmarks. 
\end{abstract}

\section{Introduction}
\label{sec:intro}

Action recognition has seen significant advances in terms of overall accuracy in recent years.
Most existing methods~\cite{Inflated3D_CVPR17,threeD_TPAMI13,TSN_TPAMI18,NonLocalNN_CVPR18,S3D_G_ECCV18}~treat this problem as a generic classification problem, of which the only difference from ImageNet~\cite{ImageNet_CVPR09}~ classification is that the input is now a video frame sequence. 
Thus, numerous efforts have been devoted to leveraging the temporal information.
However, unlike objects in ImageNet, which are usually centered and occupy the majority of pixels, human activities are complex concepts. They involve many factors such as body movement, temporal structure, and human-object interaction. 

\begin{figure}
    \begin{center}
    \includegraphics[width=1.\linewidth]{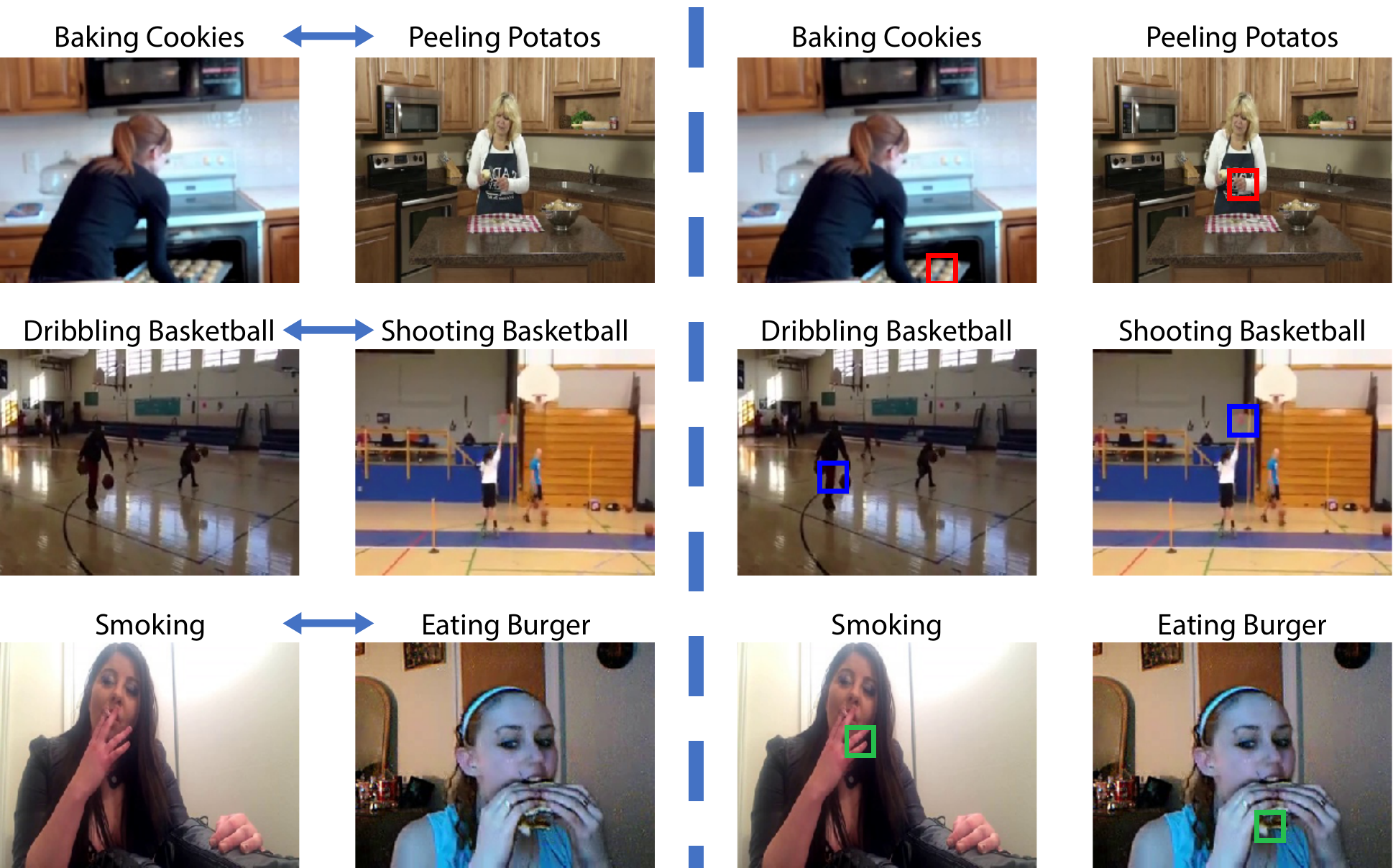}
    \end{center}
    \caption{\it Action recognition is a fine-grained recognition problem. \textbf{Left}: we illustrate samples from class pairs that are easily confused by a state-of-the-art method~\cite{Inflated3D_CVPR17}. This confusion is due to these actions being visually extremely similar and they can only be distinguished by fine-grained information. In (b) we show the classification output of the proposed approach on these classes, as well the highly activated regions with fine-grained information, such as objects (red), motion pattern (blue), object texture (green).\vspace{-2mm}}\vspace{-2mm}
    \label{fig:teaser}
\end{figure}

An example of this complexity can be found in action classes like ``eating a burger'' and ``eating a hot-dog''. They both depict eating something, and have a similar body motion pattern. To correctly distinguish them, one has to focus on the object the person is interacting with. A even more difficult situation happens when we try to distinguish ``picking up a phone'' versus ``picking up a hot-dog'', where even the shape of the objects are similar.
Another example is for ``dribbling basketball'' and ``shooting basketball'', where the objects and scenes are the same. The major cue for distinguishing them is the difference in the body motion pattern.

This complex nature of human activities dictates that rough modeling of the visual or temporal features will lead to confusion between many classes that share similar factors.
A common example of rough modeling is the widely used classifier head of one global average pooling and one linear classifier in action recognition models~\cite{Inflated3D_CVPR17,TSN_TPAMI18}, a setup typical of image object recognition models~\cite{Resnet_CVPR16} on ImageNet~\cite{ImageNet_CVPR09}.
However, as illustrated in fig.~\ref{fig:teaser}, even a state-of-the-art CNN architecture, when learned with this setup, fails to distinguish two classes when they share similar factors. In this case both classes can only be distinguished by capturing fine-grained patterns.

In this work, we propose to tackle the complexity of human action classification by promoting the importance of analyzing finer details.
In fact, a multitude of works in fine-grained recognition have been dedicated to solving similar problems on images, like distinguishing bird species~\cite{Welinder2010CUB}, car models~\cite{yang2015CompCar,Kraus2013Car} and plants~\cite{Horn2017INaturalist}.
Taking inspiration from a recent fine-grained recognition work~\cite{DiscriminativeFilterBank_CVPR18}, we propose a novel design to improve the final classification stage of action recognition models. Its main advantage is its ability to better extract and utilize the fine-grained information of human activities for classification, which are otherwise not well preserved by the global average pooling mechanism alone.
In particular, we propose to use three classification branches. The {\it first} branch is commonly used global average pooling classification head. The {\it second} and the {\it third} branches share a set of convolutions, spatial upsampling and max-pooling layers to help surface the fine-grained information of activities, and differ in terms of the classifier used. The three branches are trained jointly in an end-to-end manner. This new design is compatible with most of current state-of-the-art action recognition models, \emph{e.g.}~\cite{NonLocalNN_CVPR18,Trajectory_NIPS18} and can be applied to both 2D and 3D CNN-based action recognition methods (sec.~\ref{sec:exp:sota}).

We evaluate the proposed approach on two major large-scale action classification benchmarks: Kinetics-400~\cite{Kinetics} (400 classes) and Something-Something-V1~\cite{Something_Something_ICCV17} (174 classes).
Our results show that models built with our simple approach surpass many state-of-the-art methods on both benchmarks.
Furthermore, we also provide a detailed ablation studies and visualizations (sec.~\ref{sec:ablationstudy}) to demonstrate the effectiveness of our approach. Our discoveries suggest that the proposed approach does indeed help distinguishing similar classes that are otherwise confused by the baseline models, which leads to the improved overall accuracy.

\begin{figure*}
    \includegraphics[width=1.0\textwidth]{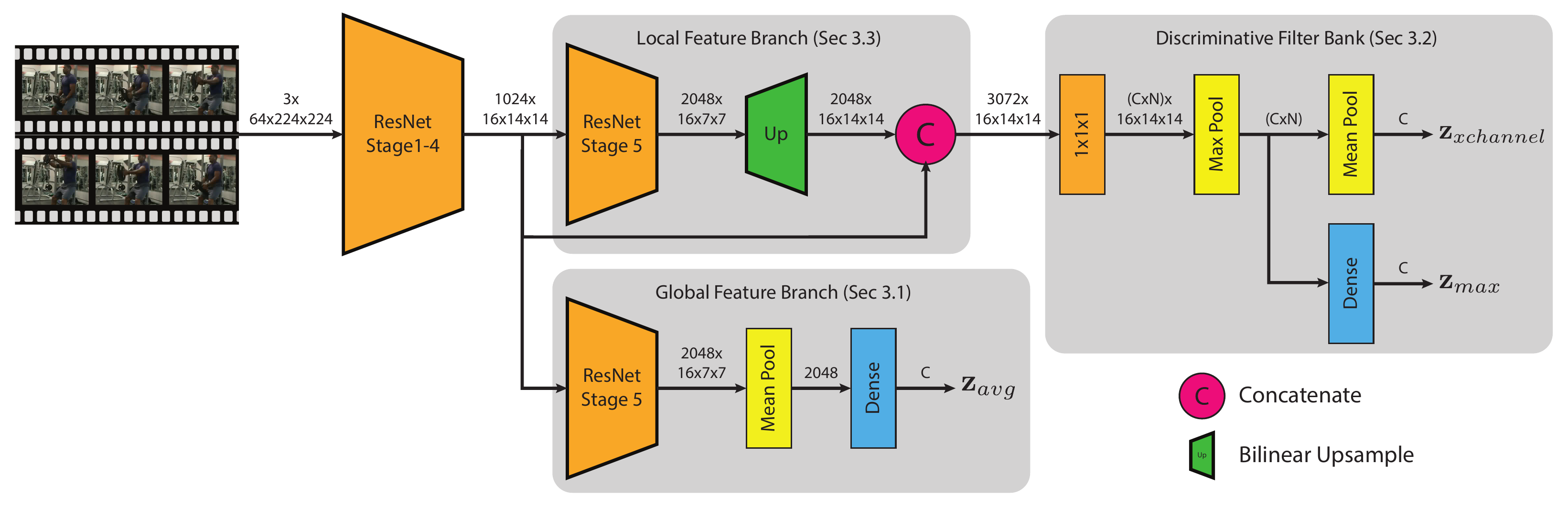}
    \caption{ \it Architecture diagram of our proposed approach. We illustrate the design with a 3D ResNet that takes 64 frames as input, but the overall design generalizes to both 2D and 3D architectures with an arbitrary number of input frames. The global feature branch (sec. \ref{sec:back-bone}) functions as our baseline. Our proposed approach improves upon this baseline with a bank of discriminative filters (sec. \ref{sec:filter_banks}) that specialize on localized cues and a local feature extraction branch (sec. \ref{sec:up}) that produces feature maps tuned to be sensitive to local patterns.}
    \vspace{-2mm}
    \label{fig:system}
\end{figure*}

\section{Related Work}
\paragraph{Action recognition in videos.}
Action recognition in the deep learning era has been successfully tackled with 2D~\cite{TwoStream_NIPS14, TSN_TPAMI18} and 3D CNNs~\cite{Inflated3D_CVPR17,Multifiber_ECCV18,threeD_TPAMI13, Tran2015C3D, R2plus1d_CVPR18,NonLocalNN_CVPR18,S3D_G_ECCV18}. 
Most existing works focus on modeling of motion and temporal structures. 
In~\cite{TwoStream_NIPS14} the optical flow CNN is introduced to model short-term motion patterns. TSN~\cite{TSN_TPAMI18} models long-range temporal structures using a sparse segment sampling in the whole video during training. 3D CNN based models~\cite{Inflated3D_CVPR17,threeD_TPAMI13, Tran2015C3D,NonLocalNN_CVPR18} tackle the temporal modeling using the added dimension of the convolution on the temporal axis, in the hope that the models will learn the hierarchical motion patters as in the image space. Several recent works have started to decouple the spatial and temporal convolution in 3D CNNs to achieve more explicit temporal modeling~\cite{Multifiber_ECCV18,R2plus1d_CVPR18,S3D_G_ECCV18}. In \cite{Trajectory_NIPS18,Yan2018STGCN} the temporal modeling is further improved by tracking feature points or body joints over time. 

Most of these methods treat action recognition as a video classification problem. 
These works tend to focus on how motion is captured by the networks and largely ignore what makes the actions unique. 
In this work, we provide insights specific to the nature of the action recognition problem itself, showing how it requires an increased sensitivity to finer details. Different from the methods above, our work is explicitly designed for fine-grained action classification. In particular, the proposed approach is inspired by recent advances in the fine-grained recognition literature, such as~\cite{DiscriminativeFilterBank_CVPR18}. 
We hope that this work will help draw the attention of the community on understanding generic action classes as a fine-grained recognition problem.

\vspace{-2mm}
\paragraph{Fine-grained action understanding.}
Understanding human activities as a fine-grained recognition problem has been explored for some domain specific tasks~\cite{Patterson2005FGAction,Singh2016Multi}.
For example, some works have been proposed for hand-gesture recognition~\cite{Fernando2015Darwin,Lea2016Segmental,Rohrbach2016Recognizing}, daily life activity recognition~\cite{Rohrbach2012Cooking} and sports understanding~\cite{Efros2003Distance,Tsunoda2017Football,Chen2011Baseball,Assfalg2002HMM}. All these works build ad hoc solutions specific to the action domain they are addressing. Instead, we present a solution for generic action recognition and show that this can also be treated as a fine-grained recognition problem and that it can benefit from learning fine-grained information. 

\vspace{-2mm}
\paragraph{Fine-grained object recognition.} Different from common object categories such as those in ImageNet~\cite{ImageNet_CVPR09}, this field cares for objects that look visually very similar and that can only be differentiated by learning their finer details. Some examples including distinguishing bird~\cite{Welinder2010CUB} and plant~\cite{Horn2017INaturalist} species and recognizing different car models~\cite{Kraus2013Car,yang2015CompCar}.
We refer to Zhao et. al~\cite{Zhao2017FGSurvey} for an interesting and complete survey on this topic. 
Here we just remark the importance of learning the visual details of these fine-grained classes. Some works achieve this with various pooling techniques~\cite{Bilinear_TPAMI17,Factorized_Bilinear_ICCV17}; some use part-based approaches~\cite{Zhang2014PRCNN,zhang2016cvpr}; and others use attention mechanisms~\cite{Zhao2017Attention,Zheng2017MAttention}.
More recently, Wang et. al~\cite{DiscriminativeFilterBank_CVPR18} proposed to use a set of $1 \times 1$ convolution layers as a discriminative filter bank and use a spatial max pooling to find the location of the fine-grained information.
Our method takes inspiration from this method and extends it to the task of video action recognition. We emphasize the importance of finding the fine-grained details in the spatio-temporal domain with high resolutions features.

\section{Methodology}
In this work we enrich the last layers of a classic action recognition network with three branches (fig.~\ref{fig:system}). We preserve the original global pooling branch as it has been shown to carry important discriminative clip-level information across all activities (sec.~\ref{sec:back-bone}). At the same time, we propose two more branches to respond to very localized structures (sec.~\ref{sec:filter_banks}). Our intuition is that in order to learn fine-grained details, the network needs per-class local discriminative classifiers from which it can pull unique signatures ($\mathbf{z}_{xchannel}$) and correlations across similar classes ($\mathbf{z}_{max}$). Finally, we propose to decouple the feature representation for the global pool branch and for the fine-grained branches to improve feature diversity (sec.~\ref{sec:up}).

\subsection{Global Branch: Baseline Network}
\label{sec:back-bone}

Our baseline method follows the recent approaches in action recognition~\cite{Inflated3D_CVPR17, TSN_TPAMI18} and it consists of a classification backbone encoder followed by the standard global average pooling and a linear classifier.  We denote this classification output as $\mathbf{z}_{avg}$. This pooling method is highly effective in capturing large contextual cues from across the video sequence, as it aggregates information from the entire spatial-temporal volume of the video via average pooling. It is designed to capture the gist of the full video sequence and, as such, it is quite effective at separating actions that take place in different scenes, like dancing vs playing golf. However, this pooling methodology limits the networks ability to focus on a particular local, spatial-temporal region as the global average forces the classifier to consider all parts of the video at once. 

\subsection{Discriminative Filter Bank}
\label{sec:filter_banks}
To allow the network to focus on key local regions of the video we introduce a set of local discriminative classifiers. Taking inspiration from \cite{DiscriminativeFilterBank_CVPR18}, we model these classifiers as filters. Each of them specializes on a particular local cue that is important for a specific action class. 
The discriminative filters are implemented as $N\cdot C$, $1\times1$ (for 2D) or $1\times1\times1$ (for 3D) convolutions followed by global max pooling over the feature volume to compute the highest activation value of each discriminative classifier. Here $C$ is the number of classes (400 for Kinetics-400 and 174 for Something-Something-V1), while $N$ is a hyper-parameter indicating how many discriminative classifiers are associated to each class (in this work we set $N=5$, as we did not observe any substantial improvement with a larger value). The max response from each filter for a given class is averaged to produce a final classification prediction. This produces a classification prediction that is highly localized, as the response comes from just $N$ locations in the feature volume. Wang \etal \cite{DiscriminativeFilterBank_CVPR18} coined this classification method cross-channel pooling, as responses are pooled in blocks of $N$ across the channel feature dimension. We adopt this nomenclature and denote this classifier output as $\mathbf{z}_{xchannel}$. We apply the standard softmax cross-entropy loss on this output and denote this loss as $\mathcal{L}_{xchannel}$. This loss directly encourages each filter to specialize on its class as their outputs are directly aggregated to produce the classification prediction.

This classifier is an aggregate of the $N$ classifiers specialized for that particular class but does not include any of the $N\cdot(C-1)$ other discriminative filters. To allow each class to benefit from all filters, we add a dense layer to the discriminative filter output after the maxpool. We denote this classifier output as $\mathbf{z}_{max}$ and apply a softmax cross-entropy loss denoted as $\mathcal{L}_{max}$. This classifier draws from local activation across weak classifiers from all classes, thus it can be thought of as a middle ground between the highly localized $\mathbf{z}_{xchannel}$ and the global $\mathbf{z}_{avg}$ 

Finally, we combine the three classifier outputs ($\mathbf{z}_{avg}$, $\mathbf{z}_{xchannel}$ and $\mathbf{z}_{max}$) into a single prediction ($\mathbf{z}_{comb}$) through simple summation, pass it through another softmax layer and compute a combined loss $\mathcal{L}_{comb}$. The final loss used in training is just a summation of all factors:

\begin{equation}
\mathcal{L} = \mathcal{L}_{comb} + \mathcal{L}_{avg} + \mathcal{L}_{max} + \mathcal{L}_{xchannel}
\end{equation}

\vspace{5mm}
We apply an auxiliary loss to each individual classification output ($\mathbf{z}_{avg}$, $\mathbf{z}_{xchannel}$ and $\mathbf{z}_{max}$) to force each to learn in isolation. All results reported in the experimental section use the aggregate classifier ($\mathbf{z}_{comb}$).

\subsection{Local Detail Preserving Feature Branch}
\label{sec:up}

While our discriminative filters provide fine-grain local cues, they still operate on the same feature volume as the average pooled classifier ($\mathbf{z}_{avg}$). This has two issues: one, the feature volume has been down-sampled to such a high degree that the filters cannot learn the finer details; and two, this feature volume is shared between the average pooled ($\mathbf{z}_{avg}$) and the discriminative filters ($\mathbf{z}_{xchannel}$ and $\mathbf{z}_{max}$) classifiers, meaning that it cannot specialize for either task.

In order to overcome these issues and improve feature diversity, we propose to branch the last stage of our backbone and use one branch for our average pooling classifier (global branch) and the other (local branch) for our discriminative filters. This allows the global branch to specialize on context while the local branch can specialize on finer details. 

Next, as we seek sensitivity to discriminative finer details, we add a bilinear upsampling operation in the local branch in charge of computing the discriminative classifiers (fig.~\ref{fig:system}) and add a skip connection from the features from stage 4. These modules provide a specialized and detailed feature volume for the discriminative filters, further enriching the information that the fine-grain classifiers can learn.

\section{Experimental Setting}
\label{sec:exp_setting}
\subsection{Datasets}
\label{sec:datasets}

We experiment on the two largest datasets for action recognition: Kinetics-400~\cite{Kinetics} and Something-Something-V1~\cite{Something_Something_ICCV17}. {\bf Kinetics-400} consists of 400 actions. Its videos are from Youtube and most of them are 10 seconds long. The training set consists of around 240,000 videos and the validation set of around 19,800 videos, well balanced across all 400 classes. The test set labels are withheld for challenges, so it is a standard practice to report performance on the validation set. Kinetics-400 is an excellent dataset for evaluation thanks to its very large scale nature, its large intra-class variability and the extensive set of action classes. 
{\bf Something-Something-V1} consists of 174 actions and it contains around 110,000 videos. These videos are shorter on average than those of Kinetics-400 and their duration typically spans from 2 to 6 seconds. %

One interesting difference between these two datasets is that, on Kinetics-400, temporal information and temporal ordering of frames are not very important and a single 2D RGB CNN already achieves competitive results compared to a much more complex 3D architecture; on the other hand, temporal information is essential for Something-Something-V1 and a simple 2D RGB CNN achieves much lower performance than its 3D counterpart. This is due to the different nature of the actions in these datasets: while they are very specific on Kinetics-400 (e.g., 'building  cabinet'), they are relatively generic on Something-Something-V1 (e.g., 'plugging something into something'). By experimenting on these diverse datasets, we show that our approach is generic and suitable to different action domains.

\subsection{Implementation Details}
\label{sec:imp_dets}

\textit{Approaches.} We experiment with two of the best performing backbones for our system: 2D TSN \cite{TSN_TPAMI18} and inflated 3D \cite{Inflated3D_CVPR17} networks. Briefly, {\bf2D TSN networks} treat each frame as a separate image. The TSN sampling method divides an input video evenly into segments (we used 3 segments in this work) and samples 1 snippet (a single RGB frame for this work) per segment randomly during training. This ensures that a diverse set of frames are observed during training. Finally, a consensus function aggregates predictions from multiple frames to produce a single prediction per video clip, on which the standard soft-max cross entropy loss is applied. In this work we use the average consensus function, as it has been shown to be effective. As these networks operate on each frame independently, we use the models provided by Gluon~\cite{chen15arxiv}, which is pre-trainined on ImageNet~\cite{ImageNet_CVPR09}.
Rather than processing each frame independently, {\bf inflated 3D networks} operate on a set of frames as a unit, convolving in both the spatial and temporal dimensions. These 3D networks are typically modeled after 2D architectures, converting 2D convolutions into 3D ones (i.e. a $3\times 3$ convolution becomes a $3\times 3\times 3$ convolution). We follow the common practice to initialize the network weights by "inflating" \cite{Inflated3D_CVPR17} weights from a 2D network trained on ImageNet. We follow the 3D implementation of Wang et. al~\cite{NonLocalNN_CVPR18}, which down-samples the temporal resolution by 4 in the initial convolutions at the beginning of stage 1, via strided convolutions. From that point on the temporal dimension remains fixed. 3D networks produce a spatial temporal feature tensor as illustrated in Fig. \ref{fig:system}.

\vspace{2mm}
\textit{Network architecture.} We use the ResNet backbone~\cite{Resnet_CVPR16} in all our experiments. For all network configurations, instead of using stride 2 to down-sample the spatial resolution in the initial $1\times1$ convolution ($1\times1\times1$ for 3D) of the first Bottleneck block, we use it in the $3\times3$ convolution ($3\times3\times3$) of the block instead.

\vspace{2mm}
\textit{Network Initialization.} We initialize our networks with the weights of a model pre-trained on the ImageNet dataset~\cite{ImageNet_CVPR09} as described above. 
The local stage 5 branch of the network does not have a corresponding pre-trained set of weights as it is not part of the standard ResNet architecture.  Two obvious choices to initialize the local branch are random initialization or use the stage 5 weights from ImageNet pre-training. We conducted preliminary experiments and found that using a random initialization for the local stage 5 branch weights gave consistently better results. We use this initialization for all results presented in this work. This is intuitive as we want the local branch to specialize for local cues, while the pre-trained ImageNet weights are already specialized for global context and are prone to remain close to that local minimum.

\textit{Optimization parameters.} We use an SGD optimizer and a starting learning rate of 0.01, which we reduce three times by factor 10 during training. We set weight decay to $10e-5$, momentum to $0.9$ and dropout to $0.5$. Our batch size is as large as permitted by the hardware and it changes with the depth of the model, the approach used (i.e., 2D or 3D) and with the number of frames of the video. For example, when using 16 frames, ResNet152 and a 3D CNN, we can only use a batch of 8. Moreover, when using 64 frames, we use mixed precision (FP16). This allows us to use larger batch sizes than with FP32.

\textit{Input size and data augmentation.} We resize all videos so the short edge is $256$ pixels. We keep the temporal resolution (FPS) the same as the source files. During training, we augment the training examples with horizontal flipping, random resizing and random cropping of $224\times224$ pixels.

\textit{Test-time settings.} We follow the standard procedures employed in the literature when comparing to other state-of-the-art methods. For the 2D models, we use 20 regularly-sampled segments at test time (a segment consists of only a frame in our case) and perform oversampling, which consists on extracting 5 crops and their flips for each segment. This results in 200 forward passes for each video. All these outputs are then averaged to obtain the final prediction. For the 3D networks, we use 10 segments. Instead of resizing each frame to a pre-defined input size, we run fully convolutional inference and average the predictions. Finally, we flip horizontally every other segment. While this is the standard testing protocol for Kinetics-400, on Something-Something-V1 we follow the standard practice to use a single segment for testing.

\begin{table}
\centering
\resizebox{0.95\columnwidth}{!}{
\begin{tabular}{|l|c|c|c|c|}
\cline{2-5}
\multicolumn{1}{c}{} & \multicolumn{2}{|c|}{2D} & \multicolumn{2}{c|}{3D} \\
\hline
Method & Top-1 & Top-5 & Top-1 & Top-5 \\
\hline
GB (Baseline) sec. \ref{sec:back-bone} &  69.6 & 88.3 & 66.8 & 86.0 \\
GB + DF sec. \ref{sec:filter_banks}& 70.9 & 89.5 &  68.0 & 86.9 \\
GB + DF + LB sec. \ref{sec:up}& 71.2 & 89.3 & 68.8 & 87.3 \\
\hline
\end{tabular}}
\vspace{2mm}
\caption{\it Results of the different components of our model on the Kinetics-400 dataset. Our baseline, which consists of a single global branch (GB) is consistently outperformed by our discriminative filters (DF) and add our specialized local branch (LB) compliments these filters, pushing performance even higher.  Performance is computed using the training-time setting. Thus, 2D models use 3 segments while 3D models use one 16-frame segment. %
\vspace{-7mm}}
\label{tab:3d_abl}
\end{table}

\section{Analysis of our system} 
\label{sec:ablationstudy}

In this section we experiment with the proposed approach on the Kinetics-400 dataset. This dataset provides a challenging benchmark for our analysis as it consists of a large set of 400 classes. First, we present an ablation study on the components of our approach (sec.~\ref{sec:exp:val}). Then, we look at what actions our model is helping and hurting the most (sec.~\ref{sec:exp:actions}). Finally, we examine qualitative results from our discriminative filter banks by visualizing their max response (sec.~\ref{sec:exp:qualitative}). We follow the standard convention and report results in terms of Top-1 and Top-5 accuracy.

\subsection{Ablation study}%
\label{sec:exp:val}
We investigate how the various components of our approach contribute to its final performance. We conduct this ablation study with a simpler inference setting than that described in sec.~\ref{sec:imp_dets}. We sample 3 segments with our 2D network and only 1 16-frames segment with our 3D network, as it is computationally prohibitive to train and test for each model variation. Results are reported in table \ref{tab:3d_abl}. The top row reports the results of our global branch ({\bf GB}) baseline, which is the standard action recognition approach (sec.~\ref{sec:back-bone}). Adding our discriminative filter bank ({\bf DF}) to it (sec.~\ref{sec:filter_banks}) consistently improves performance across all models. Finally, further enriching our model with a specialized local branch ({\bf LB} sec.~\ref{sec:up}) achieves the best performance. This shows that each component of our approach is important to improve the baseline performance. Moreover, the results show that this improvement is consistent on both 2D and 3D based networks and it shows that our approach is also generic to different action recognition baselines.

\subsection{How does performance change across actions?} \label{sec:exp:actions}
In this section we investigate how our model performs on different action classes and compare the results of the baseline 3D CNN network with those of our full approach.

\begin{table}
\centering
\resizebox{0.95\columnwidth}{!}{
\begin{tabular}{|l l | l l|}
\hline
\multicolumn{4}{|c|}{Meta-categories}\\\hline
\multicolumn{2}{|c|}{\it Significant improvement} & \multicolumn{2}{|c|}{\it Marginal improvement} \\
\hline
Waxing & +5.5 & Interact w/ animals & +1.0 \\
Swimming & +5.1 & Makeup & +0.9 \\
Cooking & +3.0 & Watersports & +0.5\\
Hair & +3.0 & Gymnastic & +0.4 \\
Using Tools & +2.9 & Athletics-Jumping & +0.4\\
Eating \& Drinking & +2.6 & Raquet-Batsports & +0.3\\
\hline
\end{tabular}}
\vspace{2mm}
\caption{\it Top-1 accuracy improvement brought by our full 3D approach over the baseline 3D network.\vspace{-3mm}}
\label{tab:metacategories}
\end{table}

\begin{figure*}[t]
    \begin{center}
    \includegraphics[width=1.0\textwidth]{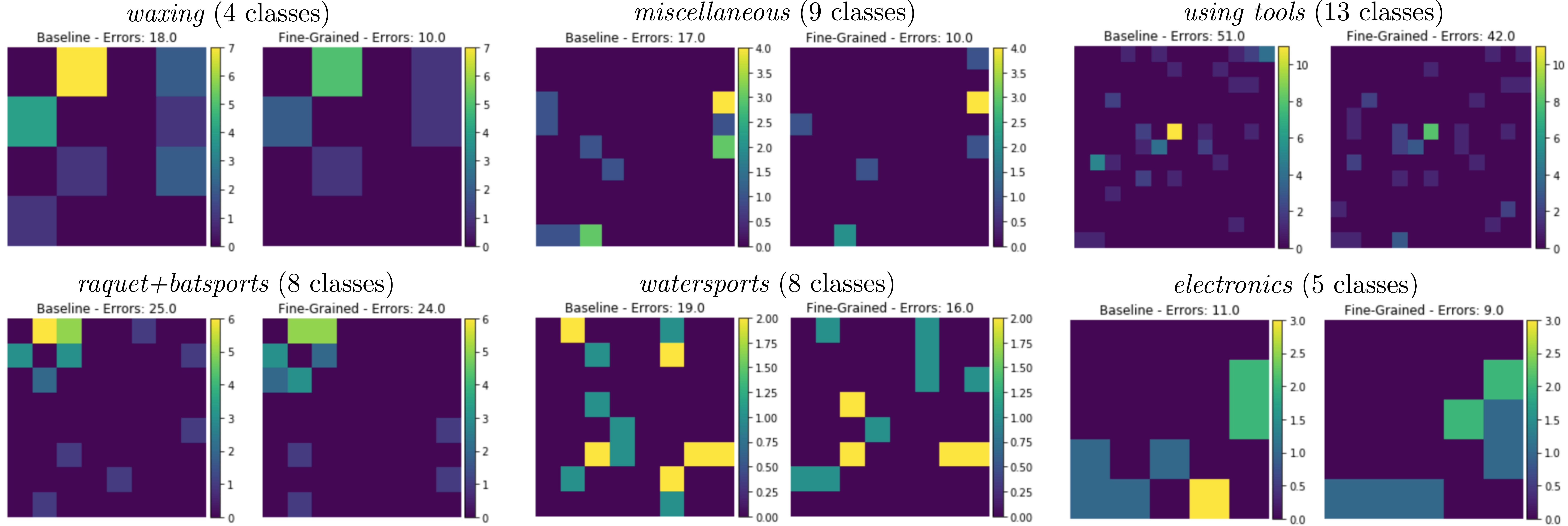}
    \end{center}
    \vspace{-3mm}
    \caption{\it \small Confusion matrices of 6 meta-categories. Our model (Fine-Grained) significantly improves Top-1 accuracy over the Baseline for the meta-categories in the top row, but only marginally for those in the bottom. Nevertheless, these confusion matrices show that our approach is much better at separating the actions within a meta-category, especially when these are visually similar. \vspace{-3mm}}
    \label{fig:cm}
\end{figure*}

\begin{figure*}
    \begin{center}
    \includegraphics[width=.49\linewidth]{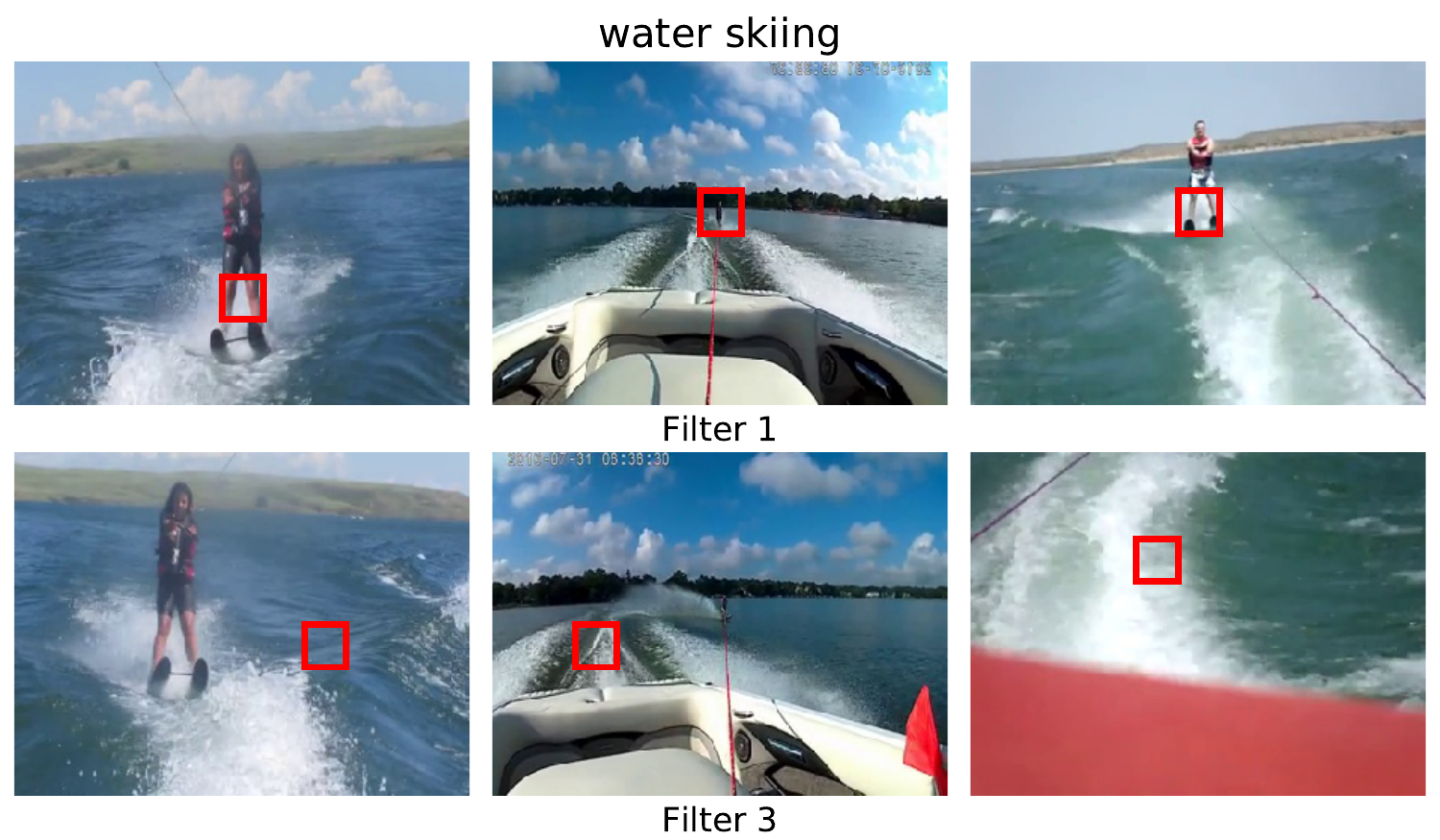}
    \includegraphics[width=.49\linewidth]{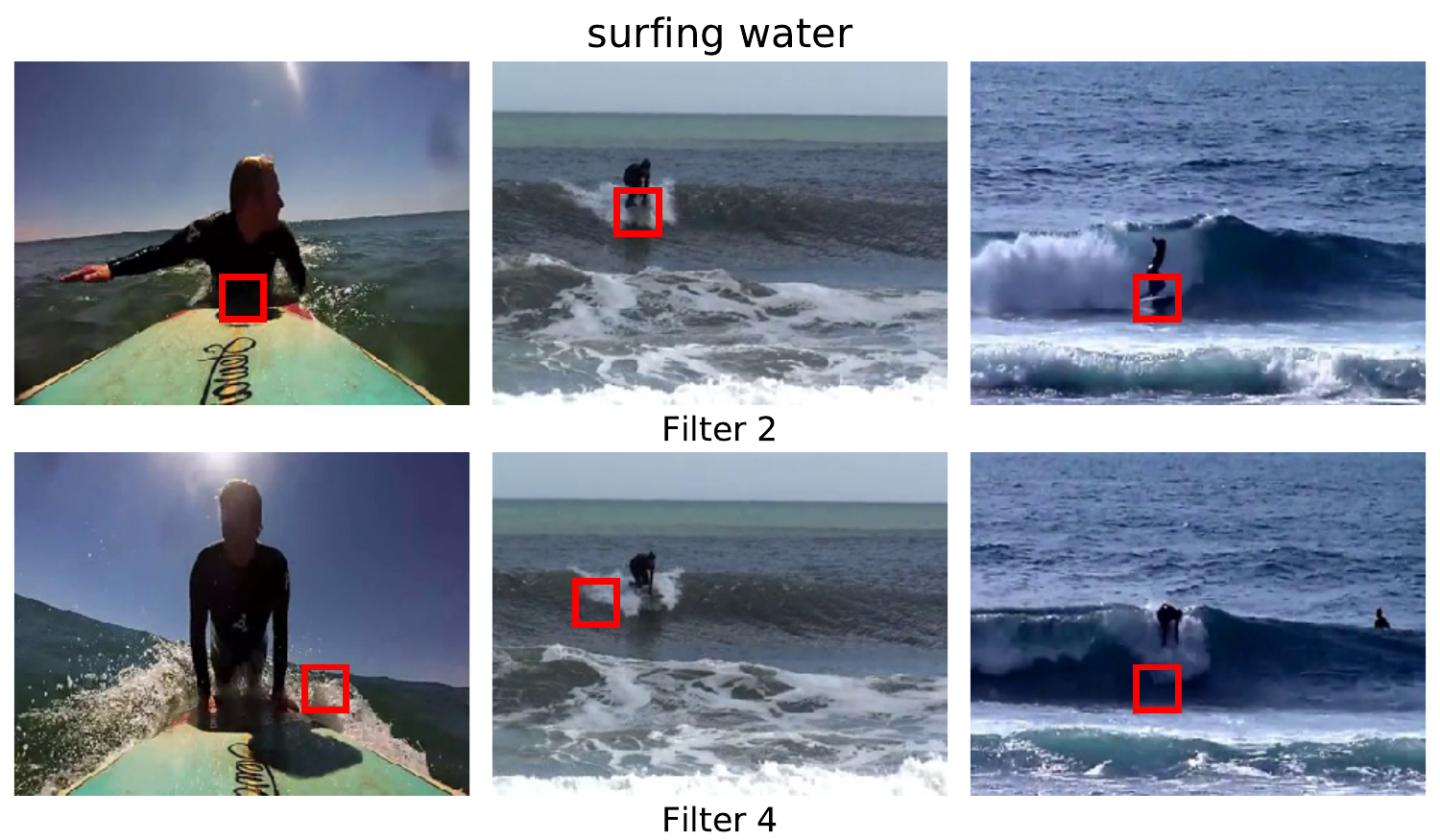}
    \end{center}
    \vspace{-4mm}
    \caption{\it Examples of 2 discriminative filters from two classes that roughly capture the same concepts. The top row shows three examples of the max response from filters that capture the person and object being ridden for the water skiing and surfing water classes. The second row shows filters that capture the texture of the water on the wake or wave in the video. Both of these cues combine to help the final classifier differentiate between these challenging classes.\vspace{-2mm}}
    \label{fig:qual_water_sports}
\end{figure*}

\begin{figure}
    \begin{center}
    \vspace{-3mm}
    \includegraphics[width=1.0\linewidth]{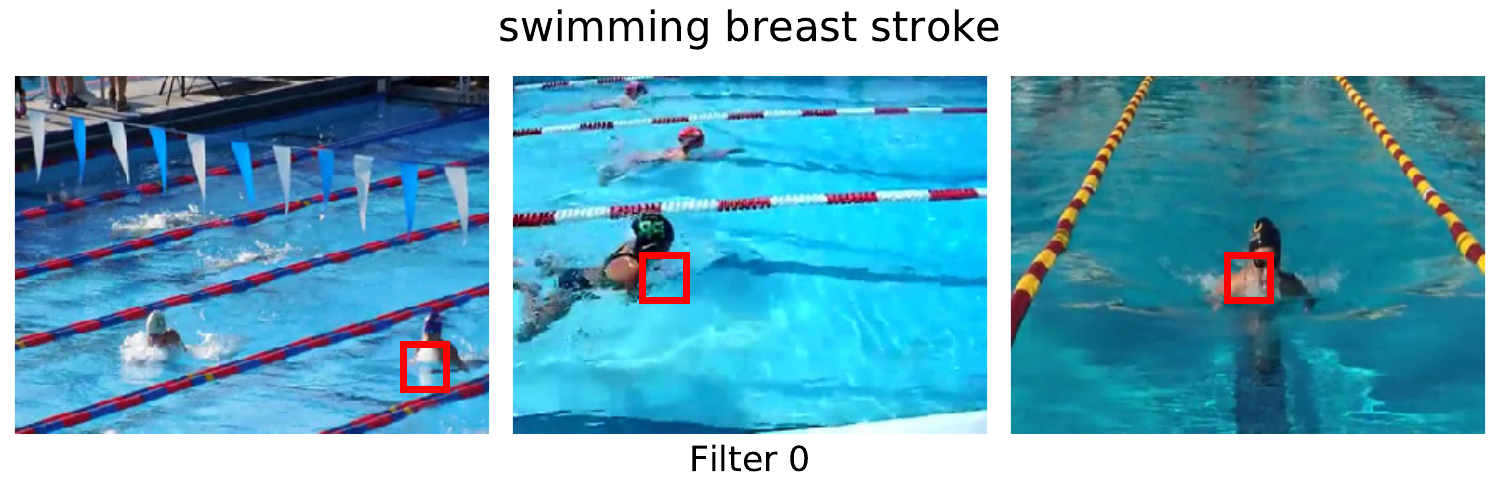}
    \includegraphics[width=1.0\linewidth]{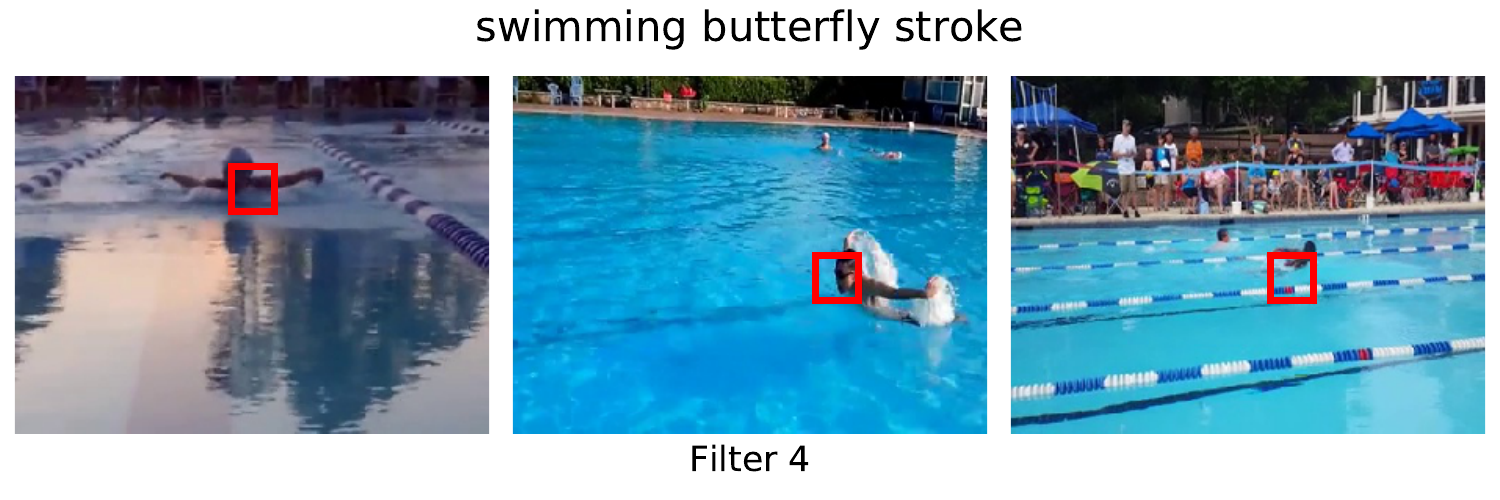}
    \includegraphics[width=1.0\linewidth]{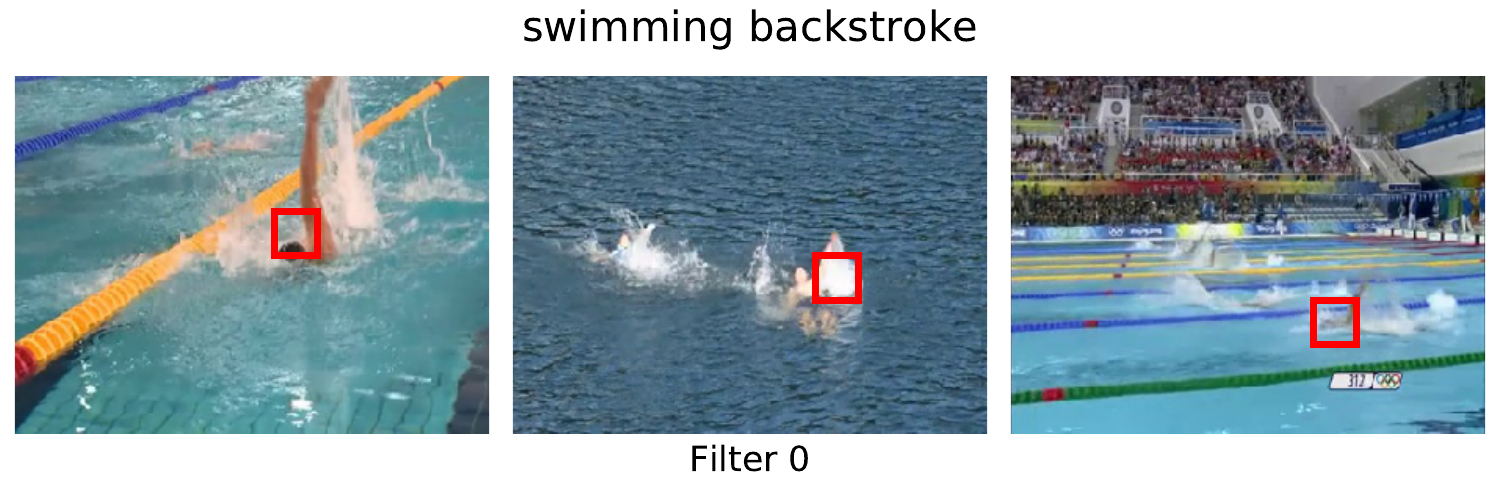}
    \end{center}
    \vspace{-6mm}
    \caption{\it Example maximal responses of a discriminative filter on three challenging swimming classes. Notice that each filter consistently fires on the frame and location of the swimmer in a canonical pose for that stroke. These filters are robust to variations in view-point and scenes. \vspace{-4mm}}
    \label{fig:qual_swimming}
\end{figure}

First, we look at the improvement of our model on high level meta-categories. These are defined by the Kinetics-400 dataset and were originally generated by manually clustering the 400 classes into 38 parent classes, each one containing semantically similar actions \cite{Kinetics}. We compute the Top-1 accuracy of a meta-category as the average Top-1 accuracy of all its child actions. In table~\ref{tab:metacategories}, we show some of the meta-categories for which we observed the largest (left) and the smallest (right) improvement. These results highlight some interesting facts. First, we observe a substantial improvement on meta-categories that contain very fine-grained actions. For example, {\it waxing} consists of classes like ``waxing back'', ``waxing chest'', ``waxing legs'', etc., which picture the same exact action, but performed on different body parts. Our approach is able to focus its attention on the relevant and discriminative regions of the videos (fig.~\ref{fig:qual_water_sports},~\ref{fig:qual_swimming}) and improve the performance on these fine-grained actions considerably (+5.5 on average for {\it waxing}). Similar trends can be observed for {\it swimming} (backstroke, breast stroke, and butterfly stroke) and several other meta-categories (table~\ref{tab:metacategories}, left). 

Interestingly, our models do not bring a similar improvement on some sport-related meta-categories, like {\it watersports}, {\it gymnastic}, {\it athletics-jumping} and {\it raquet-batsports}. This is because these meta-categories contain actions that are not necessarily similar with each other and for which the baseline model is already accurate. For example, ``playing tennis'', ``playing cricket'' and ``playing badminton'' can be easily differentiated by looking at their background scenes (i.e., the courts). Nevertheless, it is worth noticing that even on these meta-categories that are by construction unlikely to benefit from our fine-grained design, our approach still slightly improves over the baseline. This shows that our approach is robust and, overall preferable over the baseline.

\begin{table}
\centering
\resizebox{0.95\columnwidth}{!}{
\begin{tabular}{|l | l | c | c|}
\hline
\multirow{2}{*}{Action} & is most & \multicolumn{2}{c|}{\# confused videos}\\\cline{3-4}
 &   confused w/ & baseline & ours\\\hline
Bending metal & Welding & 4 & 1\\
Mopping floor & Cleaning floor & 8 & 4\\
Peeling potatoes & Baking cookies & 6 & 2\\
Plastering & Laying bricks & 4 & 0\\
Swing dancing & Salsa dancing & 8 & 4\\
Brushing hair & Fixing hair & 7 & 2\\ 
Waxing legs & Shaving legs & 4 & 1\\\hline
Getting a haircut & Shaving head & 1 & 7\\
Stretching legs & Yoga & 4 & 8\\
Strumming guitar & Playing guitar & 6 & 13\\
\hline
\end{tabular}}
\vspace{2mm}
\caption{\it Pairs of actions most confused by the baseline and our approach, along with the number of videos each model confuses.\vspace{-4mm}}
\label{tab:confusedLabels}
\end{table}

Next, we go beyond the accuracy of a meta-category and investigate its individual actions. We analyze the confusion matrices between the actions of each meta-category (fig.~\ref{fig:cm}). These visualizations clearly show that our approach is able to produce sparser confusion matrices compared to the baseline, further verifying that our approach is more appropriate to distinguish visually similar classes. For example, let's consider the meta-category {\it watersports}. Table~\ref{tab:metacategories} shows that our approach outperforms the baseline by a tiny Top-1 accuracy improvement of 0.5. Nevertheless, their confusion matrices (fig.~\ref{fig:cm}, mid-bottom) show that our fine-grained approach is capable of separating the classes within this meta-category much better than the baseline (e.g., the baseline confuses ``surfing water'' with ``waterskiing'' - cell (7,5) - but our approach differentiates them well).  

Finally, in table~\ref{tab:confusedLabels}-top we list some of the actions that are most confused by the baseline and for which our approach is more accurate. Again, we can appreciate how our approach can better separate fine-grained actions, like ``mopping floor'' and ``cleaning floor'': while the baseline wrongly predicts ``cleaning floor'' on 8 out of the 40 validation videos of ``mopping floor'', our fine-grained approach only makes 4 mistakes. Moreover, in table~\ref{tab:confusedLabels}-bottom we list actions for which our approach is more confused. While these results may seem surprising at first, they are easily explainable. These actions are connected to each other and they are in a parent-child relationship (e.g., ``shaving head'' is a type of ``haircut'') or they co-occur (e.g., ``stretching legs'' is a common pose in ``yoga''). While our model tries to learn fine-grained details of these actions, it gets confused by the ambiguity in their definition and the missing annotations, which lead to larger mistakes.

\subsection{Qualitative analysis}
\label{sec:exp:qualitative}
\vspace{-2mm}
Next, we examine qualitative examples of how our discriminate filters behave. To this end, we take activations from across an entire video of a single discriminative filter, which is trained to have a high response for a specific class. We then find the spatial-temporal location where this filter has its maximal response and draw a bounding box around the area in the frame that corresponds to that response. First, in fig.~\ref{fig:qual_swimming}, we show a single filter maximal response on three test videos for each of three swimming classes (breast stroke, backstroke, and butterfly). Notice how each filter learns to fire on the precise frame and location of the corresponding stroke's canonical pose and is robust to view-point and scene variations. We further explore this phenomenon in fig.~\ref{fig:qual_water_sports}, where we visualize two filters for two often confused classes (water skiing and surfing water). Here we find two filters for each class that roughly fire on the same types of local cues. The top row of fig. \ref{fig:qual_water_sports} shows responses for filters that focus on the person and object they are riding (water skis or surf board), while the bottom row fires on the specific wave characteristics of the filter's corresponding class. The dense classifier ($\mathbf{z}_{max}$) is then able to take advantage of these cues when making the final prediction. Finally, in our teaser figure (fig. \ref{fig:teaser}) we show more examples of how our filters are able to localize and disambiguate objects, motions, and textures.
\begin{table*}
\begin{center}
\begin{tabular}{|c|l|c|c|c|c|}
\cline{2-6}
\multicolumn{1}{c|}{} & Method & Modality & Top-1 & Top-5 & Pre-trained \\
\hline
\multirow{4}{*}{2D} & Two Stream (NIPS14) \cite{TwoStream_NIPS14}             & RGB + Flow & 65.6& -- & ImageNet \\ %
& TSN Two Stream (TPAMI18) \cite{TSN_TPAMI18}           & RGB + Flow & 73.9        & 91.1          & ImageNet \\
& TSN One Stream (our impl.)                                           & RGB            & 73.4      & 90.4        & ImageNet \\\cline{2-6}
& \cellcolor[gray]{0.9}Our approach                                                                          & \cellcolor[gray]{0.9}RGB            & \cellcolor[gray]{0.9}\bf 74.3      & \cellcolor[gray]{0.9}\bf 91.4        & \cellcolor[gray]{0.9}ImageNet \\

\hline
\multirow{9}{*}{3D} & I3D (CVPR'17) \cite{Inflated3D_CVPR17}                & RGB           & 71.1 & 89.3 & ImageNet \\
& I3D (CVPR'17) \cite{Inflated3D_CVPR17}                & RGB+Flow  & 74.2 & 91.3 & ImageNet \\
& R(2+1)D (CVPR'18) \cite{R2plus1d_CVPR18}           & RGB          & 74.3 & 91.4 & Sports-1M\\
& R(2+1)D (CVPR'18) \cite{R2plus1d_CVPR18}           & RGB+Flow & 75.4 & 91.9 & Sports-1M\\
& Multi-Fiber (ECCV'18) \cite{Multifiber_ECCV18}        & RGB           & 72.8 & 90.4 & None \\
& S3D (ECCV'18) \cite{S3D_G_ECCV18}                     & RGB          & 72.2 & 90.6 & ImageNet \\
& S3D-G (ECCV'18) \cite{S3D_G_ECCV18}                  & RGB          & 74.7 & 93.4 & ImageNet\\
& Non-local NN (CVPR'18) \cite{NonLocalNN_CVPR18} & RGB          & 77.7 & 93.3 & ImageNet \\\cline{2-6}
& \cellcolor[gray]{0.9}Our approach & \cellcolor[gray]{0.9}RGB & \cellcolor[gray]{0.9}\bf 78.8 & \cellcolor[gray]{0.9}\bf 93.6 & \cellcolor[gray]{0.9}ImageNet \\
\hline
\end{tabular}%
\vspace{-2mm}
\end{center}
\caption{\it Comparison with state-of-the-art methods in the literature on the Kinetics-400 dataset.\vspace{-2mm}}
\label{tab:sota_kinetics}
\end{table*}

\begin{table*}
\begin{center}
\begin{tabular}{|l|c|c|c|c|}
\hline
Method & Backbone & Top-1 & Top-5 & Pre-trained \\
\hline
3D-CNN \cite{Something_Something_ICCV17} (ICCV'17) & BN-Inception & 11.5 & 29.7 & Sports-1M  \\
MultiScale TRN \cite{TemporalRelationalReasoning_ECCV18} (ECCV'18) & BN-Inception & 34.4 & -- & ImageNet \\
ECO lite \cite{Eco_ECCV18} (ECCV'18) & BN-Inception+ResNet18 &  46.4 & -- & Kinetics-400 \\
I3D + GCN \cite{SpaceTimeGraphs_ECCV18} (ECCV'18) & ResNet50 & 43.3 & 75.1 & Kinetics-400 \\
Non-local I3D + GCN \cite{SpaceTimeGraphs_ECCV18} (ECCV'18) & ResNet50 & 46.1 & 76.8 & Kinetics-400 \\
TrajectoryNet \cite{Trajectory_NIPS18} (NIPS'18) & ResNet18 & 44.0 & -- & ImageNet \\
TrajectoryNet \cite{Trajectory_NIPS18} (NIPS'18) & ResNet18 & 47.8 & -- & Kinetics-400 \\
\hline
Our baseline (GB, sec.~\ref{sec:back-bone}) & ResNet18 & 42.3  & 72.3 & ImageNet \\
Our approach & ResNet18 & 45.0 & 74.8 & ImageNet \\
Our approach & ResNet50 & 50.1 & 79.5 & ImageNet  \\
\cellcolor[gray]{0.9}Our approach & \cellcolor[gray]{0.9} ResNet152 & \cellcolor[gray]{0.9}\bf 53.4 & \cellcolor[gray]{0.9}\bf 81.8 & \cellcolor[gray]{0.9}ImageNet \\
\hline
\end{tabular}
\vspace{-2mm}
\end{center}
\caption{\it Comparison with state-of-the-art methods in the literature on the Something-Something-V1 dataset.\vspace{-5mm}}
\label{tab:sota_something_something}
\end{table*}

\section{Comparison with the state-of-the-art} \label{sec:exp:sota}
In this section, we compare our method against the state-of-the-art in the literature. We use the train and test settings reported in sec.~\ref{sec:imp_dets} and report Top-1 and Top-5 accuracy.

\noindent {\bf Kinetics-400 (table~\ref{tab:sota_kinetics}).} We compare against 2D and 3D approaches. Our models achieve state-of-the-art performance on both scenarios. 
For 2D, enriching the TSN RGB stream with our fine-grained components improves both Top-1 and Top-5 accuracy by 1 point. Interestingly, our model also achieves slightly higher performance than the TSN two stream model, while remaining computationally much more efficient (Flow is extremely slow to compute and requires a forward pass through an entire second network). 
For 3D CNNS, we improve state-of-the-art Top-1 accuracy (1 point) and runtime. The previous best performing approach (Non-local Neural Networks) employs a more computationally intensive version of ResNet that has additional convolutions after most of its blocks. Instead, our approach uses a standard ResNet model that doesn't incur into this overhead. %

\noindent {\bf Something-Something-V1 (table~\ref{tab:sota_something_something}).}
We compare against 3D CNN approaches which train on RBG only. Unfortunately, the literature reports results on Something-Something-V1 using different backbone architectures, which makes a direct comparison difficult, as some architectures are trivially better than others. In order to present a fair comparison, we state the backbones used by the different approaches in our table.
We make the following observations. 
First, our approach outperforms our baseline model considerably, as it improves its Top-1 accuracy from 42.3 to 45.0 using Resnet18. Second, our approach with a ResNet18 backbone pre-trained on ImageNet improves over the previous state-of-the art (with the same settings) by $1\%$ in Top-1 accuracy (45.0 vs 44.0, TrajectoryNet~\cite{Trajectory_NIPS18}). Third, we further improve our performance by training on deeper backbones and substantially increase Top-1 accuracy by $8.4\%$  with ResNet152 backbone instead of ResNet18. Also note that TrajectoryNet improves its Top-1 accuracy by $3.8\%$ by pre-training on the Kinetics-400 dataset. While we have not tried it, we expect a similar improvement, which can further boost our performance.

Our state-of-the-art results on these two datasets validate the strength of our technique and highlight the importance of modelling action recognition as a fine-grained problem.

\section{Conclusions}
We showed that the performance of action recognition can be pushed to a new state-of-the-art by improving the sensitivity of the network to finer details. Our approach only changes the later stages of the network, thus not adding significant computational cost. It is also compatible with other methods that focus on either adding representational capacity to the backbone, or improving its computational efficiency. We achieved state-of-the-art performance on two major large-scale action recognition benchmark datasets. Moreover, this improvement is shown to generalize across different backbones, and for both 2D and 3D networks. Given these results, we hope that this work will bring attention to a previously neglected aspect of action recognition, i.e., how to enable the network to represent and learn the finer details in human activities.

{\small
\bibliographystyle{ieee_fullname}
\bibliography{action_recognition}
}

\end{document}